\documentclass[fleqn,10pt]{wlscirep}
\usepackage[utf8]{inputenc}
\usepackage[T1]{fontenc}
\title{Simultaneous lesion and brain segmentation in Multiple Sclerosis using deep neural networks }

\usepackage{changes}
\definechangesauthor[name={R M}, color=red]{rm}
\author[1, *]{Richard McKinley}
\author[1]{Rik Wepfer}
\author[1]{Fabian Aschwanden}
\author[1]{Lorenz Grunder} 
\author[1]{Raphaela Muri}
\author[1]{Christian Rummel}
\author[2]{Rajeev Verma} 
\author[3]{Christian Weisstanner} 
\author[4]{Mauricio Reyes}
\author[5]{Anke Salmen}
\author[5]{Andrew Chan} 
\author[1]{Franca Wagner}
\author[1]{Roland Wiest}

\affil[1]{Support Center for Advanced Neuroimaging, University Institute for Diagnostic and Interventional Neuroradiology, Inselspital, Bern University Hospital, Switzerland}
\affil[2]{Department of Neuroradiology, Spital Tiefenau, Switzerland}
\affil[3]{Medizinisch Radiologisches Institut, Zurich, Switzerland}
\affil[4]{Artorg Centre, University of Bern, Switzerland}
\affil[5]{University Clinic for Neurology,  Inselspital, Bern University Hospital, Switzerland}

\affil[*]{Corresponding author, richard.mckinley@insel.ch}

\begin{abstract}
Segmentation of white matter lesions and deep grey matter structures is an important task in the quantification of magnetic resonance imaging in multiple sclerosis.  In this paper we explore segmentation solutions based on convolutional neural networks (CNNs) for providing fast, reliable segmentations of lesions and grey-matter structures in multi-modal MR imaging, and the performance of these methods when applied to out-of-centre data.

We trained two state-of-the-art fully convolutional CNN architectures on the 2016 MSSEG training dataset, which was annotated by seven independent human raters: a reference implementation of a 3D Unet, and a more recently proposed  3D-to-2D architecture (DeepSCAN).  We then retrained those methods on a larger dataset from a single centre, with and without labels for other brain structures.  We  quantified changes in performance owing to dataset shift, and changes in performance by adding the additional brain-structure labels.  We also compared performance with freely available reference methods.  

Both fully-convolutional CNN methods substantially outperform other approaches in the literature when trained and evaluated in cross-validation on the MSSEG dataset, showing agreement with human raters in the range of human inter-rater variability.  Both architectures showed drops in performance when trained on single-centre data and tested on the MSSEG dataset. When trained with the addition of weak anatomical labels derived from Freesurfer, the performance of the 3D Unet degraded, while the performance of the DeepSCAN net improved.  Overall, the DeepSCAN network predicting both lesion and anatomical labels was the best-performing network examined.
\end{abstract}
\begin{document}

\flushbottom
\maketitle

\thispagestyle{empty}

\section*{Introduction}

MR-based imaging biomarkers have been integral diagnostic criteria and disease progression markers for multiple sclerosis for more than 20 years \cite{McFarland92}.  Established biomarkers for MS include baseline MRI lesion count and temporal evolution of lesion load, as well as the integrity of the blood brain barrier, signatures of axonal damage, and atrophy of cortical and subcortical grey matter \cite{Wattjes2015}.  These measures of progression are routinely assessed by expert human readers  \cite{Graber2011BiomarkersOD}.   Manual volumetric measurements are not, however, part of clinical routine, owing to the time and work required to produce an accurate segmentation.   There is a considerable variation in lesion quantification between expert readers with different backgrounds (e.g. neuroradiologists vs. neuroimmunologists), and also substantial intra-rater variation ~\cite{Altay2013}. Automated methods are by their nature repeatable, and an increasing number of publications have investigated machine learning methods to provide lesion quantification. Recently, improvements in hardware and software have enabled fast training and application of convolutional neural networks (CNNs), leading to a number of deep learning approaches to lesion segmentation.  In the 2016 MSSEG lesion segmentation challenge at MICCAI, for example, CNN-based approaches won both arms of the challenge \cite{mckinley2016, Valverde2017, Commowick2018}.  Subsequent papers have used the data arising from this challenge as a reference dataset, improving on the results achieved~ \cite{Hashemi2109}. 

Automated segmentation methods are already in regular use as a tool in imaging studies \cite{boonstra_tremor_2017,thaler_t1_2017,thaler_use_2018,marschallinger_geostatistical_2018,guo_repeatability_2019,zhang_predicting_2019}.  To train an in-house lesion segmentation tool requires both high-quality training data, voxel-level annotation by domain experts, and skills in statistical or machine-learning.  As a result, most studies employ one of a few freely available tools: for example, those cited in the previous sentence all employed the Lesion Segmentation Toolkit (LST), a plugin for SPM ~\cite{Schmidt2012}.
 This means that transferability is vital: a classifier trained on data from a single or small number of centres must be able to operate well on data from centres and scanners not seen during training, since researchers are unlikely to have access to large amounts of labelled training data.  Often CNN-based methods do not exhibit this property: performance is substantially degraded on external data, with retraining on additional labelled data being required before good performance is re-established \cite{Valverde2018}. Valverde and coworkers (2018) analysed the effect of domain adaptation on their proposed CNN-based MS lesion segmentation method, evaluating the minimum number of annotated images needed from the new domain. Domain adaptation was progressively more effective with increasing training cases, but the models still yielded a remarkably high performance on reduced training sets, such as a single training case. 
Volumetry of deep grey matter structures is also highly relevant in MS: a recent study of 1,417 MS patients showed an association between deep grey matter volume loss and clinical progression in MS \cite{Eshaghi2018}.  In that study, segmentation of the grey matter structures was not carried out directly, since the tool used (FSL-FIRST) is known to be biased by the presence of white matter lesions.  The white matter lesions were identified using LST and removed from the imaging using 'lesion filling'.  

In this paper, we study modern deep learning methods performing a simultaneous segmentation of both lesions and grey matter structures, trained on a combination of lesion segmentations produced by manual raters, and weak labels for the healthy-appearing tissue provided by an existing automated method (Freesurfer).  Deep learning has already been applied successfully to the segmentation of healthy brains \cite{Wachinger2017, Rajchl2018}.  Our hypothesis is that modern deep learning architectures can simultaneously perform the tasks of lesion and anatomy segmentation, and that these methods are robust enough when applied to data outside of the training sample to distribute and use in the context of clinical studies.

\section{Methods}

\subsection{Study design and overview}

We studied in this work the performance of CNN-based lesion segmentation methods on two datasets: a small publicly available dataset, and a larger dataset from our own centre.  The datasets belonging to the MSSEG challenge conform to the OFSEP guidelines, guaranteeing 3D FLAIR and a 3D T1 imaging~\cite{Commowick2018,COTTON2015}.  Briefly, 15 datasets were obtained from three different centres in France, each using a different scanner: the images were annotated by seven trained junior experts, and their segmentations were fused to provide a single ground truth for training purposes using LOP-STAPLE \cite{Akhondi14}.  Our internal dataset comprised 122 fully anonymized MRI datasets of patients from the Inselspital (Bern Switzerland) which form part of the MS cohort of the University of Bern. Ethical approval for the study was granted by the local ethical commission (Cantonal Ethical Commission Bern, 'MS segmentation disease monitoring', approval number 2016-02035) \added[id=rm]{and all patients gave general consent for data storage and analysis of their MRI datasets.}
We split this data into two cohorts: 90 datasets were included in the Insel90 dataset (50 used for training and 40 for validation of the classifier), and 32 patients were included in the Insel32 dataset (used in this paper for testing the classifier).   All datasets stemmed from patients who fulfilled the diagnostic criteria (revised McDonald criteria of 2010/2017) for relapsing-remitting MS ~\cite{Polman2011}.   Datasets from the Inselspital were acquired using a standardised acquisition protocol on a 3T MRI (Siemens Verio, Siemens, Erlangen, Germany).  See Table~\ref{tab:scanseq} for details of the sequences used on each scanner.

Lesion annotation of the Insel90 and Insel32 datasets was conducted by two of the authors (RWe/FA), under the supervision of an experienced neuroradiologist with more than 15 years experience in MS (RWi). Manual segmentations of the MS lesions were acquired through slice-by-slice analysis of all four sequences.  The 3DSlicer platform \cite{FEDOROV2012} was used to perform the manual lesion segmentation following a standardised protocol \cite{GarciaLorenzo2013}. Lesions were identified if appearing hyper-intense compared to the surrounding normal-appearing WM on T2w and FLAIR images, and slightly to severely hypo-intense on T1-weighted images.  
A final consensus segmentation was prepared by (RMu), supervised by an experienced neuroradiologist with more than 10 years of experience (FW) who had access to the initial segmentations. This multi-reader 'consensus' segmentation between  three raters was used for training and validation of the classifier. One of the authors (RWe) also annotated severely T1 hypo-intense 'black holes', which were added as a final label.

Silver standard segmentations of the cortical grey matter, cortical white matter, ventricles, cerebellum, thalamus, caudate, putamen, pallidum, hippocampus, amygdala, brain stem, ventral diencephalon, choroid plexus, corpus callosum and accumbens area were extracted from the segmentation (aseg.mgz) generated by applying Freesurfer 5.3 to the MDEFT acquisition. No distinction was made between left and right hemispheres.  In addition to these tissue labels, Freesurfer also generated labels for white matter hypointensities, which encompass not only lesions due to MS, but also age-related physiological changes in the periventricular regions (mild ependymal loss, subependymal gliosis and widened extracellular space: tissue labelled as white matter hypointensity but not labelled by our raters as MS lesion were assigned this label (non-lesion WMH).

\begin{table}[]
\centering
\begin{tabular}{llll}
\multicolumn{1}{c}{Dataset} & \multicolumn{1}{c}{Scanner} & \multicolumn{1}{c}{Sequences} &  \\ \cline{1-3}
\multicolumn{1}{c}{}        &                             & 3D FLAIR $0.74 \times 0.74 \times 0.7$ mm     &  \\
                            & Philips Ingenia 3T          & 3D T1w $0.74\times0.74\times0.85$ mm      &  \\
                            &                             & 2D T2w $0.74\times0.74\times0.85$ mm      &  \\ \cline{2-3}
                            &                             & 3D FLAIR $1.03\times1.03\times1.25$ mm    &  \\
MSSEG                       & Siemens Aera 1.5T           & 3D T1w $1.08\times1.08\times0.9$ mm       &  \\
                            &                             & 2D T2w $0.72\times0.72\times4$ mm         &  \\ \cline{2-3}
                            &                             & 3D FLAIR $0.5\times0.5\times1.1$ mm       &  \\
                            & Siemens Verio 3T            & 3D T1w $1\times1\times1$ mm               &  \\
                            &                             & 2D T2w $0.69\times0.69\times3$ mm         &  \\ \cline{1-3}
                            &                             & 3D FLAIR $ 0.48\times0.48\times1 $mm       &  \\
Insel                       & Siemens Verio 3T            & 3D T1w $1\times1\times1$ mm               &  \\
                            &                             & 2D T2w $0.42\times0.42\times3$ mm         & 
\end{tabular}
\caption{Scanner and sequence details  for the two datasets used in this paper}
\label{tab:scanseq}
\end{table}

The MSSEG datasets were supplied together with brain masks.   Brain masks were prepared for the 122 T1-weighted MRIs in the Insel datasets using FSL-BET with hand-tuned parameters. These brain masks were then transferred to the T2 and FLAIR images using rigid registration by FSL-FLIRT.  In addition, we trained a DeepSCAN network \cite{Mckinley2018a} to predict the BET brain-mask from T1, T2 or FLAIR imaging, to allow application of our method to new cases without relying on BET.  For the purposes of training segmentation algorithms, the data for each case (comprising T1, T2 and FLAIR imaging) was skull-stripped, rigidly registered to the FLAIR image, and resampled to 1mm isovoxels.

\added[id=rm]{In common with standard practice, before training the voxel intensities were standardized, by subtracting the mean of voxel intensities withing the brain mask, and dividing by the standard deviation of intensities within the brain mask.  No other pre-processing was applied}

\subsection{3D-to-2D network: DeepSCAN}
In a previous papers on brain tumor segmentation, and longitudinal multiple sclerosis and brain segmentation, we proposed the DeepSCAN architecture.\cite{Mckinley2018a,Mckinley2019b,McKinley2019c} DeepSCAN is a '3D-to-2D' architecture, meaning that the network takes as input a 3D volume, and predicts labels for the central 2D slice of that volume. The 3D-to-2D structure of DeepSCAN's layers combines the benefits of a fully 3D network (3D context) and a 2D network (smaller memory footprint of the input, leading to more available memory for feature maps)

\begin{figure}[htbp]
\centering
\includegraphics[width=12cm]{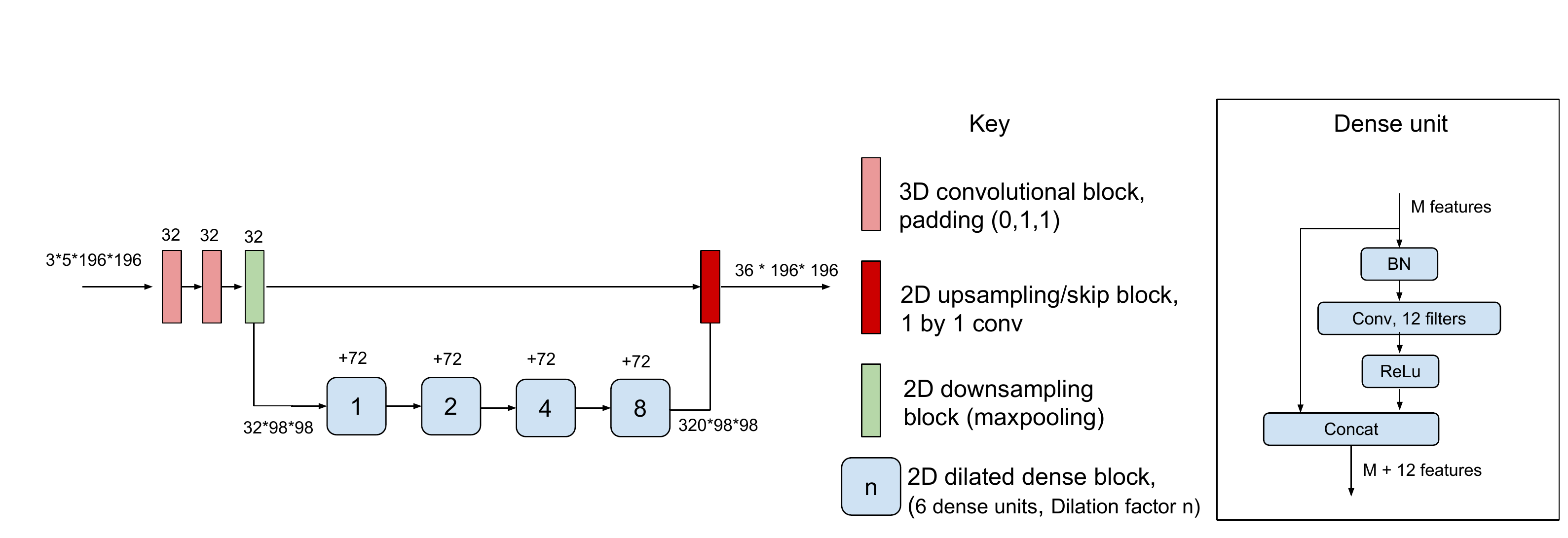}
\caption{The DeepSCAN architecture used in this paper for lesion and brain-structure segmentation\label{deepscanMS}}
\end{figure}

A DeepSCAN architecture for MS lesion and neuroanatomy segmentation is shown in Figure~\ref{deepscanMS}.  Following \cite{Mckinley2018,Mckinley2019b,McKinley2019c}, the network was trained using a combination of multi-class cross-entropy loss and a \emph{label-flip loss} for each individual tissue class which quantifies label uncertainty.  Training of the classifier was performed using the ADAM optimiser, with a cosine-annealing learning rate schedule applied over each epoch \cite{LoshchilovH16a2016}.  The batch size was two, with each 3*5*192*192 block being randomly sampled in either an axial, coronal or sagittal direction from one of the training examples.   \added[id=rm]{The model was trained for 100 epochs (where here an epoch was defined to be  2000 batches): for the first 50 epochs, a cosine-annealing learning rate schedule running from 1e-4 to 1e-7, for epochs 50-100 a cosine annealing learning rate running from 1e-5 to 1e-9. For regularization we used weight decay (1e-3).  Since the lesion class is the class of most interest, and is a minority class among voxels, training slices were selected so that each batch contained at least ten voxels of lesion tissue.} 

When applied to a new case, the classifier is applied in the axial, coronal and sagittal direction.  Minimal test-time augmentation is performed by also segmenting the volumes mirrored in the sagittal direction. An ensembling of these six outputs is then performed by averaging raw logit outputs.

\subsection{Publicly available benchmark methods: nicMSlesions and SPM Lesion Segmentation Toolbox}

Several methods for segmenting multiple sclerosis lesions are available to download as Docker containers, self contained software, or SPM modules.  We compare the results of our algorithm to a CNN-based MS lesion segmentation algorithms, nicMSlesions \cite{Valverde2017}, a patch-based CNN method intended for MS lesion segmentation available from Github (\url{https://github.com/sergivalverde/nicMSlesions/}).    We also applied algorithms from the Lesion Segmentation Toolkit  (available at \url{https://www.applied-statistics.de/lst.html}),  which are widely used in neuroimaging studies.  The toolkit provides two segmentation algorithms: a lesion prediction algorithm (LPA), operating on FLAIR imaging alone, and a lesion growth algorithm (LGA), operating on FLAIR and T1 imaging.  As recommended by the authors of the toolkit, we used our training cases to determine the best initial threshold for the LGA algorithm.  LST and nicMSlesions have both been used as benchmark methods in recent publications on MS lesion segmentation~\cite{larosa2020, cerri2020contrast}.

\subsection{3D Unet (nnUnet)}
 The above-mentioned benchmark tools for MS lesion segmentation are all based on rather outdated methodologies.  Results for 3D Unets (the prevailing architecture for 3D medical image segmentation)  on the MSSEG dataset as reported by Salehi et. al and Hashemi et al.~\cite{salehi2017,Hashemi2109} suggest that a vanilla Unet performs worse, on that dataset, than the winner of the 2016 MSSEG challenge, a 2D CNN~\cite{mckinley2016}, which is somewhat surprising.  The best performing network in that study was a 3D Unet variant which, like the DeepSCAN network, incorporates dense connections.    To ensure that our method performs at a high level we want to compare with a state-of-the-art 3D trained for comparison a version of 3D Unet: since the precise Unet architectures used in Salehi et al and Hashemi et al is not available, we use the Unet as implemented by Isensee et al. in the nnUnet framework \cite{Isensee2019}, on the MSSEG datasets, and on the Insel90 dataset.  By using the predefined settings of that framework, which have been shown in the Medical Segmentation Decathlon (\url{http://medicaldecathlon.com}) to work well on a variety of medical segmentation tasks, we can ensure that performance of the DeepSCAN network is being measured versus a strong benchmark.

\subsection{Experiments}
We first trained a DeepSCAN model, 3D Unet (nnUnet), and nicMSlesions on 50 of the Insel90 cases, using the 40 remaining cases for validation, with only the consensus lesion label as ground truth.  To compare the effect of adding additional labels, we then trained the same model architectures (apart from nicMSlesions, which is only designed to segment a single label), from scratch, with additional silver standard tissue labels as derived from Freesurfer.  On the Insel32 dataset, we assessed the performance of the models with and without additional silver standard labels.  Benchmark metrics for comparison this dataset were provided by the freely available tools LST LPA and LST LGA. \cite{Schmidt2012}.

We then applied the Unet and DeepSCAN and nicMSlesions models, trained on the Insel90 dataset, to the training data of the MSSEG dataset without retraining or tuning, to test the effectiveness of these models on multi-centre out-of-sample data.  To assess the effect of out-of-sample training, we also trained a 3D Unet and a DeepSCAN model, in cross-validation, on the MSSEG training data itself (weights for a nicMSlesions model trained on the MSSEG dataset are provided in the downloadable version).

\subsection{Performance metrics}

For performance on MS lesion segmentation, we adopted the performance metrics of the MSSEG challenge: a segmentation metric (Dice coefficients) and a lesion detection metric (lesionwise F1 score).  Definitions of the metrics used can be found in the supplementary material.  On the Insel32 dataset we compared to the consensus segmentation, while on the MSSEG dataset we compared both to the individual raters and to the LOP-STAPLE consensus.~\cite{Commowick2018, Akhondi14}

Direct evaluation of segmentation quality for the brain anatomy labels is not feasible, since the work involved in producing high quality full-brain annotations is too high: as a proxy we calculated Dice coefficients for selected brain structures on both the Insel90 and MSSEG datasets: here our goal is to determine the relative performance between these two datasets, and between the two model architectures.
\section{Results}

\begin{figure}
\centering
    \includegraphics[width=10cm,
    keepaspectratio,]{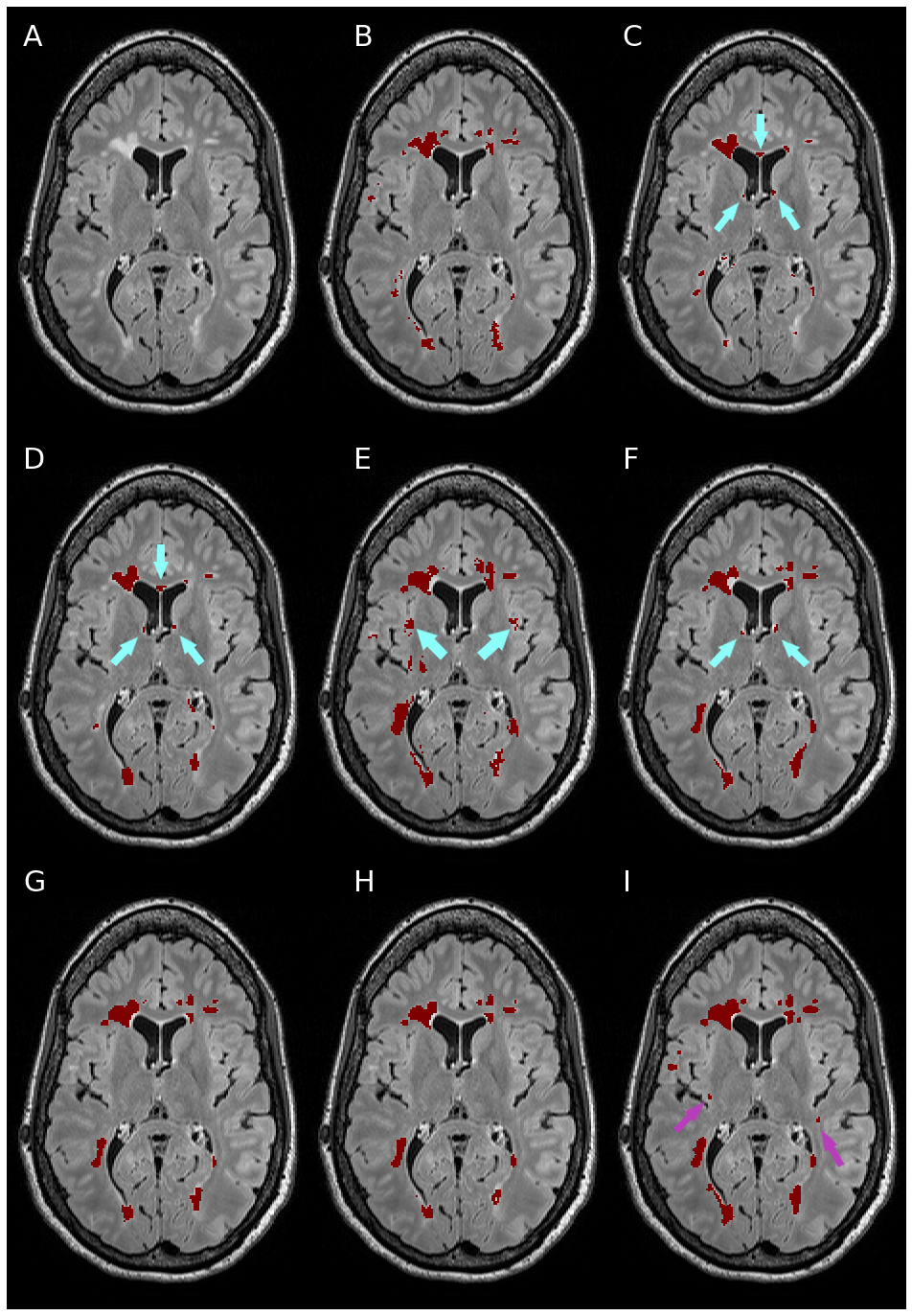}
    \caption{Example segmentations of a case from the Insel32 dataset, axial view.  (A) FLAIR image, plus segmentations from (B) manual raters, (C) LPA, (D) LGA, (E) nicMSlesions and (F) nnUnet (all labels) (G) DeepSCAN (just lesions) (H) nnUnet (just lesions) (I) DeepSCAN (all labels). Segmentations from LGA, LPA, and nnUnet (all labels) show erroneous segmentation of healthy-appearing tissue near the ventricles  (cyan arrows).  nicMSlesions labels healthy cortical grey matter as lesion tissue (cyan arrows) DeepSCAN (all labels) identified two subtle lesions (violet arrows) missed by the manual raters but subsequently confirmed by our experienced neuroradiologists.}
    \label{fig:lesion_segsb}

\end{figure}

\subsection{Test set performance on the Insel32 dataset}
Performance metrics on the Insel32 dataset (vs the consensus segmentation) are summarised in Table \ref{table:Insel_results}, along with inter-rater comparison between the two raters, and between the raters and the consensus. An example of output of the methods considered, together with the manual ground truth, can be seen in Figure~\ref{fig:lesion_segsb}.

We performed Wilcoxon signed-rank tests of the differences between the methods, with a significance level of 0.05.  All DeepSCAN and nnUnet models trained performed significantly better than LPA, LGA and nicMSlesions, according to both Dice coefficient and F1 score.
No significant difference was found between the Dice coefficients of DeepSCAN (lesions only), DeepSCAN (lesions and brain structures) and nnUnet (lesions only): these three methods were significantly better regarding Dice coefficient than nnUnet (lesions and brain structures).  Regarding lesion detection (lesion F1 score), DeepSCAN (lesions and brain structures) was significantly better than all other methods.

\subsection{Performance on MSSEG data: cross-validation vs out-of-sample performance on LOP-STAPLE consensus}
Both the 3D Unet and the DeepSCAN model achieved state-of-the-art performance on the MSSEG dataset when trained in cross-validation, against the LOP-STAPLE consensus segmentation, comfortably outperforming the previously reported best model from Hashemi et al.~\cite{Hashemi2109} (the previous best performance reported in the literature) where the best-performing method reached a Dice coefficient of 70.3 using a Densely connected Unet with a Dice loss. Indeed, performance of both DeepSCAN and 3D Unet exceeded that of the mean individual rater in terms of Dice coefficient. This suggests that both a plain 3D Unet (nnUnet) and DeepSCAN are performing at state-of-the-art levels, when trained in cross-validation on the MSSEG dataset. Both methods also substantially outperformed the publicly available nicMSlesions tool with default weights, which was trained on data including the MSSEG training dataset (and should therefore, if anything, be somewhat overfitted to this dataset).

All methods saw degradation of performance when trained on the larger single centre Insel90 dataset and tested on the MSSEG dataset, in line with previous studies.  Notably, the two fully-convolutional CNN methods performed, when trained on out-of-sample data, similarly to the nicMSlesions (patch-based) CNN trained on in-sample data.   The trend observed in the Insel90 dataset, where additional labels improved results for DeepSCAN and worsened results for 3D Unet, is present also in the MSSEG dataset.

A summary of performance for the methods tested on the MSSEG dataset, plus a summary of the inter-rater performance on this dataset, is shown in Table~\ref{tab:insel-to-msseg}.

\begin{table}[]
\centering
\begin{tabular}{l c c}

                                        & Mean Dice Coefficient & Mean Lesion F1 Score \\ \hline
Rater A vs Rater B                      & 54.8           & 0.42            \\ 
Rater A vs consensus                    & 62.7           & 0.57             \\ 
Rater B vs consensus                    & 60.3           & 0.55             \\ \hline
LST LPA*                                 & 31.9             & 0.21            \\ 
LST LGA*                                & 42.26            & 0.25            \\ \hline
nicMSlesion                             & 44.24            & 0.31                \\ 
nnUnet (lesions only)                   & 56.2             & 0.48            \\ 
DeepSCAN (lesions only)                 & 57.29            & 0.52            \\ 
nnUnet (lesions and brain structures)   & 51.99            & 0.42            \\ 
DeepSCAN (lesions and brain structures) & \textbf{60.0}    & \textbf{0.57}   \\ \hline
\end{tabular}
\caption{Segmentation and lesion detection metrics on the Insel32 dataset, together with inter-rater statistics.  Methods labelled * were not trained on the Insel90 dataset, and are provided for reference}
\label{table:Insel_results}

\end{table}

\subsection{Performance on MSSEG data versus individual raters}
The mean Dice coefficient between two individual raters on the MSSEG dataset ranged between 54 and 90 \footnote{The Dice coefficient between raters R4 and R5 is 90, an outlier result which seems to have resulted from communication between the raters, as between all the other raters the Dice coefficient is below 75. For many lesions R4 and R5 provide identical contours. See for example Figure 1 of Commowick et al.~\cite{Commowick2018}, where sub-figures (d) and (e) have identical annotations for many lesions}.  Dice coefficients for the various different versions of nnUnet and DeepSCAN can be seen in Table~\ref{tab:raters}.  As can be seen, both methods trained in Cross-validation are comfortably within the range of inter-rater Dice coefficients, while of the methods trained on the Insel90 dataset, only DeepSCAN (all labels) performs within the inter-rater range.

\subsection{Segmentation of healthy-appearing  grey matter structures on the Insel32 and MSSEG dataset}
\begin{figure}
    \centering
    \includegraphics[width=14cm,
    keepaspectratio,]{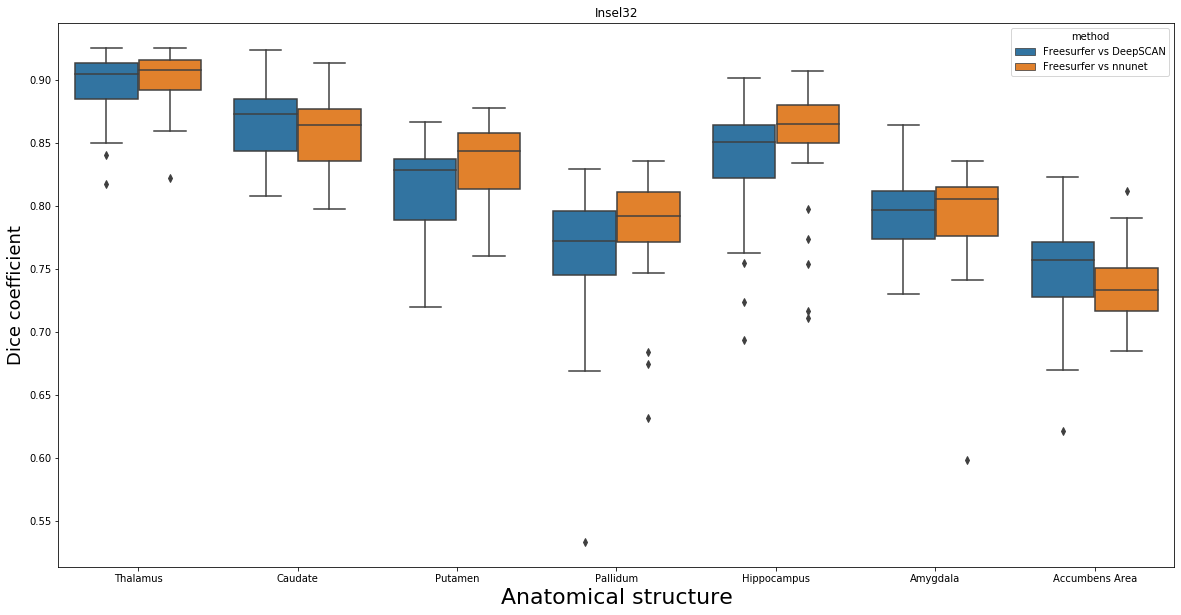}
    
    \vspace{1em}

    \includegraphics[width=14cm,
    keepaspectratio,]{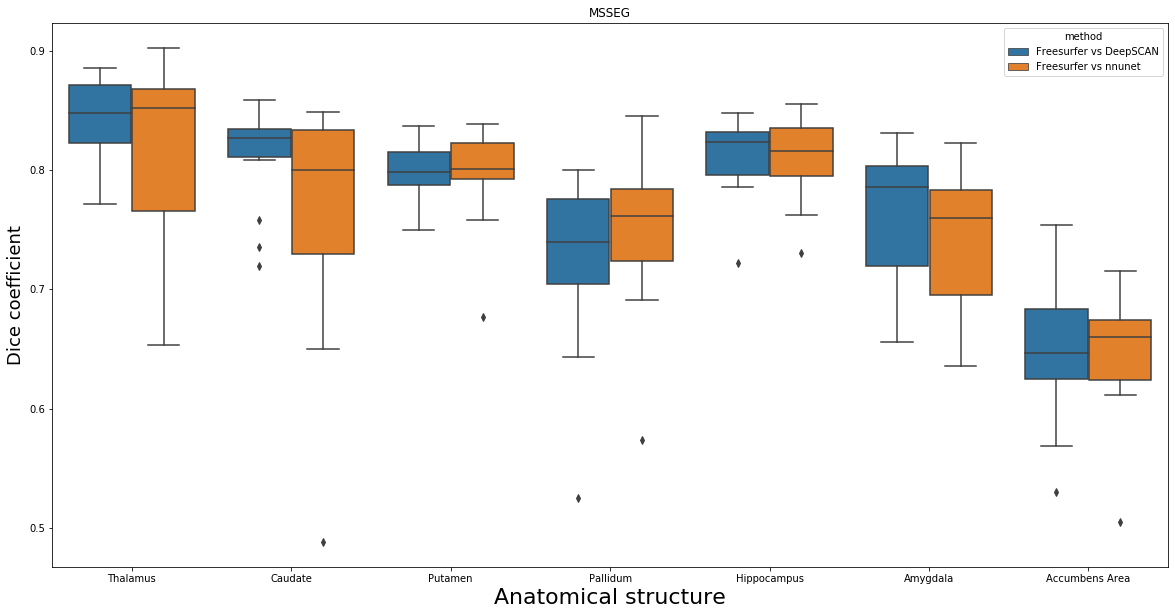}
    \caption{Boxplots of Dice coefficients for selected grey matter structures, for classifiers trained on the Insel90 dataset, and validated on the Insel32 (in sample) and MSSEG (out of sample) datasets.} 
    \label{fig:boxplots_anatomical}
\end{figure}

\begin{figure}
    \centering
    \includegraphics[width=6cm,
    height=6cm,
    keepaspectratio,]{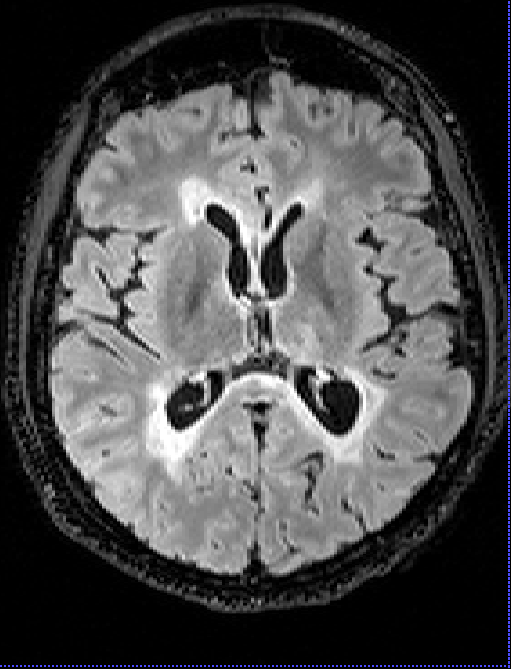}
    \includegraphics[width=6cm,
    height=6cm,
    keepaspectratio,]{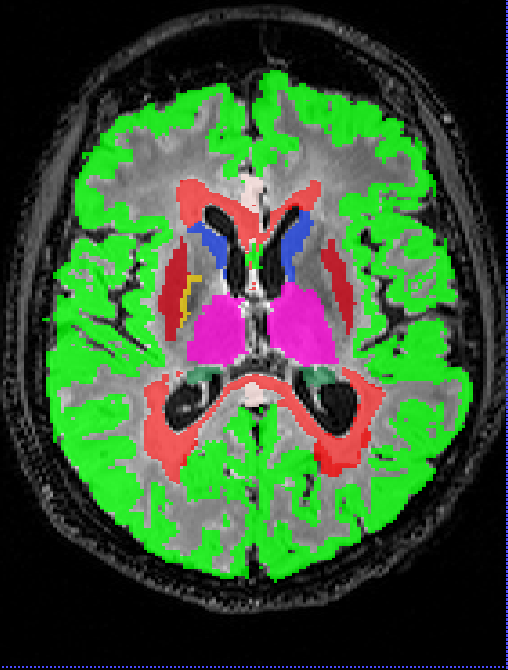}
    
    \includegraphics[width=6cm,
    height=6cm,
    keepaspectratio,]{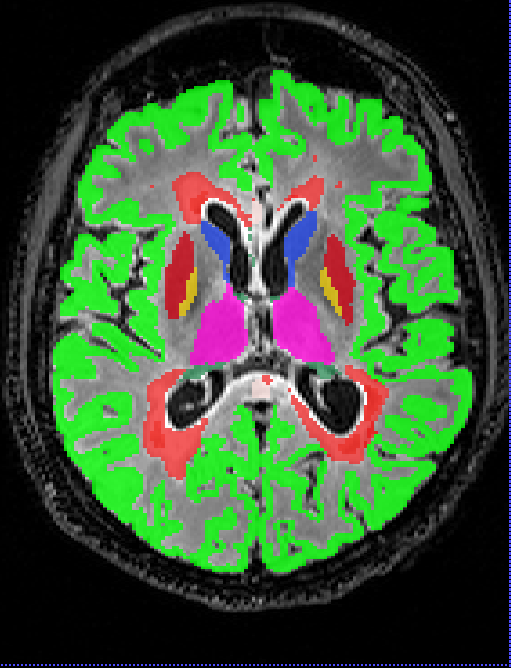}
    \includegraphics[width=6cm,
    height=6cm,
    keepaspectratio,]{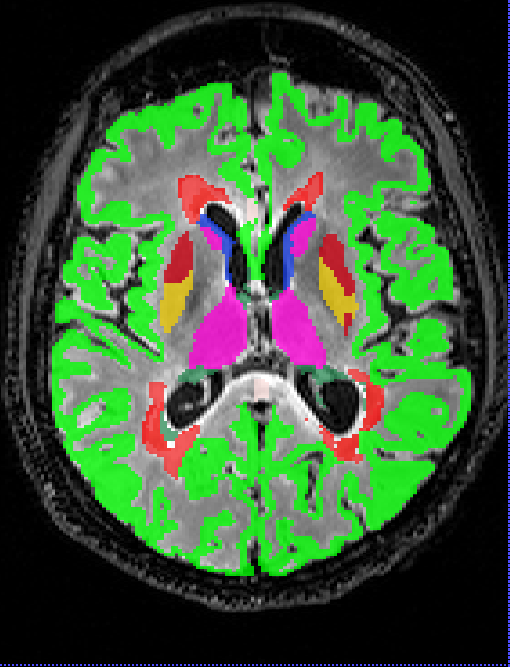}
    \caption{An axial slice of case 1016SACH from the MSSEG dataset.  Top left, FLAIR, top right, fused 'ground truth', bottom left, DeepSCAN segmentation, trained on the Insel90 dataset, bottom right, nnUnet segmentation, trained on the same Insel90 dataset.  nnUnet model incorrectly labels parts of Caudate nucleus as Thalamus, and incorrectly places boundary between Pallidum and Putamen.}
    \label{fig:full_segmentation}
\end{figure}

 Box plots of the Dice coefficient for the segmentation (DeepSCAN multi-task) of various grey matter structures, when compared with Freesurfer, can be found in Figure~\ref{fig:boxplots_anatomical}.  Both models performed quite robustly when applied to the Insel32 dataset. However, the nnUnet model can be seen to make substantial errors in the placing of structures on the MSSEG (out of sample) dataset, as can be seen in Figure~\ref{fig:full_segmentation}.

\begin{table}[]
 \centering
\begin{tabular}{ l c c c c }
                      & Training &  & Dice  & Lesion-wise  \\ 
                      & Data set & Labels &  Coefficient & F1 Score \\ \hline
Expert raters (inter-rater) & - & - & 63.1   & 0.53 \\      
Expert raters (mean vs consensus)        & - & - & 73.4             &  0.69\\ \hline          
nicMSlesions          & MSSEG & Lesions &  58.7            & 0.40 \\ 
nnUnet          & MSSEG & Lesions & 74.5             & 0.62 \\ 
DeepSCAN        & MSSEG & Lesions &  75.7    & 0.59          \\ \hline
nicMSlesions          & Insel90 & Lesions &   40.8           &  0.26 \\  
nnUnet & Insel90 & Lesions & 61.8            & 0.47            \\ 
DeepSCAN &  Insel90 & Lesions &  59.7   & 0.51   \\               
nnUnet & Insel90 & All labels &    60.5            & 0.47            \\ 
DeepSCAN &  Insel90 & All labels &  66.1   & 0.55   \\ \hline
\end{tabular}

\caption{Segmentation and lesion detection metrics against the MSSEG LOP-STAPLE consensus}
\label{tab:insel-to-msseg}
\end{table}

\begin{table}[]
\centering
\begin{tabular}{c c c c c c c c c}

  & nicMS     & DeepSCAN & Unet    & nicMS   & DeepSCAN & Unet    & DeepSCAN   & Unet       \\
 & MSSEG & MSSEG    & MSSEG   & Insel90 & Insel90  & Insel90 & Insel90    & Insel90    \\
       & lesions   & lesions  & lesions & lesions & lesions  & lesions & all labels & all labels \\ \hline 
R1            &    51.9       & 62.7     & 62.3    &    34.6     & 50.0     & 51.7    & 56.0       & 50.5       \\  
R2            &     54.0      & 62.3     & 63.3    &    36.7     & 51.0     & 51.5    & 61.0       & 50.7       \\  
R3            &     53.1      & 67.6     & 65.2    &      37.7   & 56.1     & 57.9    & 60.6       & 55.7       \\  
R4            &     51.8      & 58.4     & 60.0    &    36.0     & 48.5     & 48.8    & 59.0       & 48.4       \\ 

R5            &     53.4      & 61.3     & 61.9    &     36.5    & 50.0     & 50.7    & 60.3       & 49.9       \\  
R6            &     51.3      & 66.3     & 64.7    &     36.2    & 57.4     & 58.0    & 60.0       & 56.2       \\  
R7            &     49.9      & 68.3     & 64.6    &     34.4    & 58.1     & 60.1    & 56.9       & 57.5       \\ \hline 
Mean          &     52.2      & 63.8     & 63.1    &     36.0    & 53.0     & 54.1    & 59.11      & 52.7       \\ \hline 
\end{tabular}
\caption{Segmentation performance (Dice coefficient) versus the seven individual raters of the MSSEG challenge}
\label{tab:raters}
\end{table}

\section{Discussion}

We explored the performance of our in-house developed 3D-to-2D (DeepSCAN) CNN architecture on the task of MS lesion segmentation, both with and without simultaneous labelling of healthy-appearing structures, comparing it to a state-of-the-art implementation of a 3D Unet, and also to existing, publicly available lesion segmentation algorithms, some of which are widely used in neurological and other imaging studies. When trained and tested in cross-validation on the multi-centre MSSEG dataset, both DeepSCAN and the 3D Unet outperformed previously reported  state-of-the-art methods on te same dataset. The two algorithms achieved better segmentation agreement with the MSSEG LOP-STAPLE consensus than the mean human rater, and also broadly agreed with each other on the lesion segmentation (the mean Dice coefficient between the cross-validated Unet and DeepSCAN models was 80.5, higher than all but one pair of individual human raters).

When trained on a larger single-centre dataset, both DeepSCAN and 3D Unet outperformed publicly available lesion segmentation tools on our internal validation set.  
All three of Unet, DeepSCAN and nicMSlesions showed drops in performance (relative to the MSSEG cross-validated classifiers) on lesion segmentation and detection tasks when trained on the single-site dataset. This is consistent with previous studies~\cite{Valverde2018}; however, the fully-convolutional networks trained on the Insel data-set had out-of-sample performance  similar to the top-performing models in the original MSSEG challenge and at the lower end of inter-rater variability.  We would like to mention here in addition that it is difficult to disentangle the effects of the shift in data (between different scanners) and the shift in the concept of MS lesion between the lesion annotation protocols (for example, care was taken in the Insel dataset to avoid labelling enlarged perivascular spaces as lesion tissue, which does not seem to be the case in the MSSEG dataset).

Including weak labels for neuroanatomical structures caused a drop in performance for the 3D Unet, but a gain in performance for the DeepSCAN network: this change in performance was more pronounced when testing on out-of-sample data.  As always, when dealing with deep learning models, it is difficult to attribute a cause to this effect, but the reduced performance of the Unet on a larger set of labels could perhaps be because of the ratio between the number of labels (18) and the feature depth (30) at the highest resolution layers of the network. Compared this with a feature depth of 320 at the highest resolution for DeepSCAN.  Increasing the feature depth of the Unet while still allowing training on a single GPU would require reducing the size of the input patch (currently 128 by 128 by 128), which would undoubtedly lead to decreases in performance due to reduced context (as is the case for nicMSlesions, which operates with 11 by 11 by 11 patches).  A full study of this trade-off deserves its own subsequent analysis.

Mean Dice coefficients between the independent expert raters on the MSSEG dataset ranged between 54 and 90.  Meanwhile the agreement between DeepSCAN and the individual MSSEG raters was between 57 and 61 for the classifier trained on the Insel90 dataset, and between 58 and 68 for the classifier trained in cross-validation.  This suggests that automated classification can operate with similar inter-rater variability to human raters, and also points towards the limits of using manual segmentations to judge the performance of segmentation algorithms in MS: in spite of using three raters, supervised by two experts, the lesion annotation we used to train our 3D Unet and DeepSCAN models contained (like almost all volumetric annotations of 3D imaging datasets) errors. We were able to quantify the extent of these errors by 
 retrospectively examining the results of DeepSCAN algorithm.   Two raters (FW and LG) made a final assessment of each false positive lesion (i.e. labelled by the algorithm but not overlapping the ground truth by more than 10\%), and each false negative lesion (i.e. labelled in the ground truth but not overlapping with the segmentation of the algorithm by more than 10\%). The raters were shown each potential false positive/negative lesion, blind to whether the lesion segmentation came from the manual raters or the algorithm.  The goal was to identify what proportion of 'false positive' lesions were in fact missed by all manual raters, and what proportion of 'false negative' lesions were in fact mislabelled (i.e. were normal-appearing grey matter, choroid plexus, or the result of imaging artefacts).  Over all 32 test subjects in the Insel32 dataset, there were a total of 2701 lesions annotated  in the consensus labelling.  Of the 2701 lesions in the consensus ground truth, the multi-task DeepSCAN classifier failed to find 777.  However, 579 out of those 777 (75\%) were judged by a pair of trained neuroradiologists working together as being mislabelled (i.e. were normal-appearing grey matter, choroid plexus, or the result imaging artefacts).  Of the 2019 lesions annotated by DeepSCAN multi-task, 372 did not overlap more than 10\% with lesions in the consensus ground truth.  137 of those lesions (37\%) were subsequently judged to be genuine lesions which had been missed by the manual raters.   
 These results, plus the remaining gap in performance between the human raters and algorithms on Lesion Detection (the F1 score) suggest that future challenges should focus much more on detection measures than segmentation performance in terms of Dice.

There are several limitations of the study that we note, which may be improved in subsequent work.  Firstly, the proposed method is restricted to work on high quality data similar to that acquired in the MSSEG challenge, with 3D T1 and FLAIR imaging and 2D T2-weighted imaging.  In practice these sequences might not all be available.  However, this combination of sequences is typical of imaging available in university clinics and used in clinical studies.  This contrasts with other recent approaches which have employed MP2RAGE~\cite{larosa2020} or 7 Telsa imaging~\cite{fartaria2019}. It would, of course, be ideal to present a method which can perform at a high level on a variety of imaging contrasts.   A very recent manuscript by Cerri et al.~\cite{cerri2020contrast} proposes a generative model for MS lesion and neuroanatomy segmentation which is contrast-adaptive, allowing lesion segmentation from a variety of different contrasts.  However despite the  authors claims  that purely supervised models cannot be expected to perform well on out-of-sample data out-of-sample data, this much more sophisticated method achieved a Dice coefficient of 0.57 (vs DeepSCAN's 0.59) against the MSSEG raters, despite having been trained on a substantially larger dataset than in this study (212 MS cases from a single centre).  It would therefore be interesting to investigate the ability of purely supervised methods to cope with a variety of modalities at input, simply by data augmentation. Second, we note that our method, which uses 3D convolutions, is not designed to handle data with 2D rather than 3D resolution: nor is it able to work at a higher resolution than 1mm isovoxels.   This again limits the applicability and performance of the trained model.  Third, both the silver-standard Freesurfer-derived labels and the manual labels for lesion contained errors, limiting the reliability of the performance metrics.  Manual checking of the false/true positives allows to compensate for that, but owing to the substantial work involved in manually checking each lesion candidate, we were only able to carry  out the this analysis for one of the trained algorithms (we selected the DeepSCAN (all labels) algorithm as it was the best performing on both the Insel32 and MSSEG datasets). Furthermore, since the MSSEG dataset and Insel datasets were annotated by disjoint sets of raters, the effects of scanner- and site-variation in imaging are difficult to disentangle from systematic differences in how the lesions were annotated.   A more useful comparison for the healthy-appearing tissue labels would be between our automated tool and  expert manual labels (although these, too, will suffer from substantial inter- and intra-rater variability): alternatively, as in Cerri et al~\cite{cerri2020contrast} a comparison could be made to more robust tools, incorporating lesion filling.    Finally, we only examined cross-sectional performance:  in a companion study~\cite{Mckinley2019b}, we demonstrate the applicability of the DeepSCAN classifier for detecting changes in lesion loads, making use of the segmentation uncertainty, but it would also be of merit to validate longitudinal performance of a range of classifiers.  

In conclusion, we have presented evidence that CNN-based tools have the potential to provide segmentations of MS lesions with agreement to human raters similar to inter-rater agreement.  Depending on the architecture, these tools may also segment healthy-appearing brain tissue without losing performance on lesion segmentation.  Such a tool has the potential to improve both the treatment of MS patients and the quality of neuroimaging studies. 

\bibliography{main}

\section*{Acknowledgements}

This research was supported by the Swiss Multiple Sclerosis Society (“Automatic segmentation of MS lesions, brain volumetry and morphometry: Proposal of a diagnostic tool for disease monitoring”) and a grant from the Novartis Forschungsstiftung ("DeepSCAN-Cortex").

\section*{Author contributions statement}

R.Mc., M.R., F.W. and R.Wi. conceived the experiments,  R.Wi, F.A., F.W., C.W., R.V., A.S., and A.C. identified and provided data.  R.We, F.W., F.A., R.W., L.G. and R.Mu. provided data annotations. R.Mc. and C.R. provided implementations of the algorithms.  R.Mc., conducted the experiments and analysed the results.  R.Mc., F.W. and R.Wi composed the manuscript.  All authors reviewed the manuscript. 

\section*{Additional information}

Competing interests:

Richard McKinley, Rik Wepfer, Fabian Aschwanden, Lorenz Grunder, Raphaela Muri, CHristian Rummel, Rajeev Verma, Christian Weisstanner, Mauricio Reyes, Franca Wagner and Roland Wiest have no interests to declare.  Anke Salmen has recieved speaker honoraria and/or travel compensation
for activities with Almirall Hermal GmbH, Biogen, Merck, Novartis, Roche, and Sanofi Genzyme, none related to this work.
Andrew Chan  received honoraria for board and speaker honoraria from Actelion, Bayer, Biogen, Celgene, Merck, Novartis, Sanofi-Genzyme, Roche, Teva, all for hospital/university research funds. He has recieved reserach funds from Research funds: UCB, Biogen, Sanofi-Genzyme, and is an editor for European Journal of Neurology, and Clin Transl Neurosci.

\appendix
\section{Segmentation and detection performance metrics}

\subsection{Dice coefficient}

The \emph{Dice coefficient}, also known as the Dice-S\o{}renson coefficient or Dice Similarity score, is a standard measure of segmentation agreement.  If X and Y are two sets, the Dice coefficient is defined as 
\begin{equation}
    \mathrm{Dice}(X,Y) = \frac{2 |X \cap Y|}{|X| + |Y|}
\end{equation}

To maintain consistency with other publications, we display the Dice coefficient as a percentage (i.e. a number between 0 and 100)

\subsubsection{Lesion detection rates}
In addition to assessing the quality of voxel-by-voxel detection of lesion tissue, we also assess detection at the lesion level.  We follow the proposed lesion detection metrics of the MSSEG 2016 challenge, for reference see Commowick et al ~\cite{Commowick2018}.  

A "lesion" in a given segmentation will be defined as a connected component of the segmentation (with 18-connectivity kernel) of size $3mm^3$ or greater.  We use the notation $\mathrm{D}_X(Y)$ to stand for the detection rate of $X$-lesions in the segmentation $Y$, meaning the proportion of lesions in $X$ successfully detected by $Y$.  For these purposes, a lesion in $X$ is detected by $Y$ if:
\begin{itemize}
    \item The lesion overlaps with lesions in $Y$ by at least $\alpha \%$. 
    \item The lesions which contribute most to the detection of the lesion (those summing up to $\gamma \%$ of the overlap) do not lie more than $\beta \%$ outside the lesion.
\end{itemize}

A lesion can thus fail to be detected either by "undersegmenting" (the lesion is missed, or is detected but the corresponding lesions in $Y$ are too small), or by "oversegmenting" (the corresponding lesions in $Y$ are too large).  We adopt the values $\alpha = 10$, $\beta = 70$, $\gamma = 65$ from \cite{Commowick2018}.  The value $D_X(Y)$ is then simply the ratio of the number of lesions detected to the total number of lesions in $X$.  If $X$ is the "ground truth" and $Y$ is any other segmentation , then $D_X(Y)$ measures the proportion of true lesions which were correctly segmented, or sensitivity, of the segmentation $Y$.  In this case, $D_Y(X)$ is the proportion of lesions segmented by $Y$ which are in "true lesions",  i.e. the \emph{precision} or Positive Predictive Value.  The F1 score summarizes these two values in one statistic, and is given by the harmonic mean of $D_X(Y)$ and $D_Y(X)$:

\begin{equation}
    F1(X,Y) = 2 \frac{D_X(Y) D_Y(X)}{D_X(Y) + D_Y(X)}
\end{equation}
 
\end{document}